# Image Super-resolution via Feature-augmented Random Forest

Hailiang Li[1], Kin-Man Lam[1], Miaohui Wang[2]

[1]Department of Electronic and Information Engineering, The Hong Kong Polytechnic University
[2]College of Information Engineering, Shenzhen University, Guangdong, China

harley.li@connect.polyu.hk, enkmlam@polyu.edu.hk, mhwang@szu.edu.cn

**Abstract** — Recent random-forest (RF)-based image super-resolution approaches inherit some properties from dictionary-learning-based algorithms, but the effectiveness of the properties in RF is overlooked in the literature. In this paper, we present a novel feature-augmented random forest (FARF) for image super-resolution, where the conventional gradient-based features are augmented with gradient magnitudes and different feature recipes are formulated on different stages in an RF. The advantages of our method are that, firstly, the dictionary-learning-based features are enhanced by adding gradient magnitudes, based on the observation that the non-linear gradient magnitude are with highly discriminative property. Secondly, generalized locality-sensitive hashing (LSH) is used to replace principal component analysis (PCA) for feature dimensionality reduction and original high-dimensional features are employed, instead of the compressed ones, for the leaf-nodes' regressors, since regressors can benefit from higher dimensional features. This original-compressed coupled feature sets scheme unifies the unsupervised LSH evaluation on both image super-resolution and content-based image retrieval (CBIR). Finally, we present a generalized weighted ridge regression (GWRR) model for the leaf-nodes' regressors. Experiment results on several public benchmark datasets show that our FARF method can achieve an average gain of about 0.3 dB, compared to traditional RF-based methods. Furthermore, a fine-tuned FARF model can compare to or (in many cases) outperform some recent state-of-the-art deep-learning-based algorithms.

1. INTRODUCTION

In the past few years, random forest (RF) [3, 14] as a machine-learning tool, working via an ensemble of multiple decision trees, has been employed for efficient classification or regression problems, and applied to a large variety of computer-vision applications, such as object recognition [27], face alignment [15, 18, 21], data clustering [17], single image super-resolution (SISR) [8, 19], and so on.

The RF method, which benefits from its simple implementation of binary trees, has been widely used, and exhibits a number of merits, including (1) it works with an ensemble of multiple decision trees to express the principle that "two heads are better than one", (2) it is easy to be sped up with parallel processing technology, on both the training and inference stages, (3) it has sub-linear search complexity, because of the use of the binary tree structure, (4) the bagging strategy for feature candidates on split-nodes enable it to handle high-dimensional features and avoid over-fitting on regression, and (5) the clustering-regression scheme employs the "divide and conquer" strategy, which can tackle the classification and regression tasks with more stable performance.

The RF-based image super-resolution approach can be considered as a clustering/classification-based method, as shown in Fig. 1. But the clustering and regression problems in RF require with different discriminative features, which have not been systematically studied in existing literature.

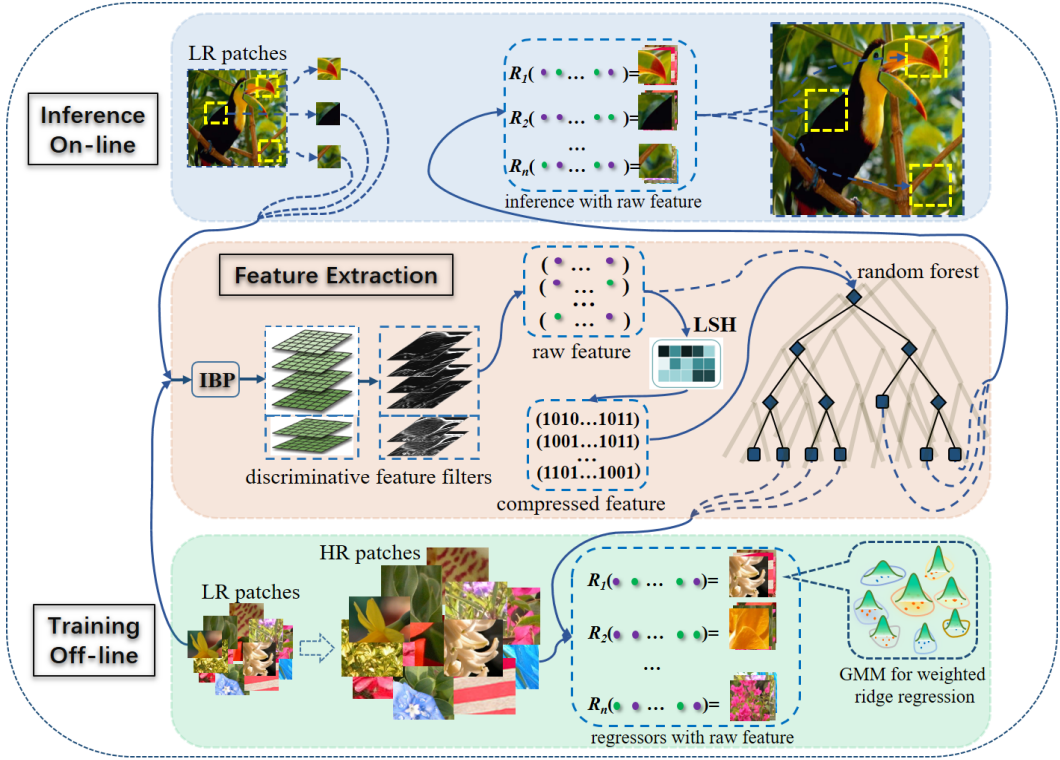

Fig. 1: An overview of the proposed FARF framework for image super-resolution.

Feature engineering has been a research hotspot for decades. Several features have been proposed for learning the mapping functions from low-resolution (LR) patches to high-resolution (HR) patches on image restoration problems. Pioneer work in [45] used a simple high-pass filter as simple as subtracting a low-pass filtered values from the input image raw values. Meanwhile, most algorithms [1, 2, 4, 5, 8] follow the approach in [28], which concatenates the first- and second- order gradients to form the features, as an inexpensive solution to approximating high-pass filtering. Since RF is used as a dictionary-learning-based tool, it inherits many properties from the conventional dictionary-learning-based algorithms on feature extraction. However, the discriminative ability of those gradient-based features for random forest has been overlooked in the literature. We found, from experiments, that augmented features based on two gradient-magnitude filters can achieve more than 0.1dB quality improvement in RF based SISR, with the same parameter setting.

In most dictionary-learning-based algorithms, principal component analysis (PCA) is used for dimensionality reduction before classification and regression processes. The impact of using PCA has also been paid less attention in the literature. PCA projection may damage the structure of features, which are originally discriminative for clustering at the split-nodes and regression at the leaf-nodes. Motivated from content-based image retrieval (CBIR) [46, 47], where the coarse-level search uses compressed features, while the fine-level search uses augmented features. Therefore, in our method, we use the original features rather than the compressed features generated by PCA as worked in [1, 2, 4, 5, 8, 28], so that more accurate regression and higher image quality improvement can be achieved. Moreover, the unsupervised locality-sensitive hashing (LSH) model, instead of PCA, is employed for feature dimensionality reduction, which can reduce the damage on the feature structure for the compressed features used on clustering at the split-nodes and thus improve the final image quality.

For regression problems at the leaf-nodes, we propose a generalized weighted ridge regression (GWRR) as an extension of the work in [1]. GWRR models are generated based on the data distributions from the leaf-nodes.

The main contribution of our method is on feature augmentation, so we call our method feature-augmented random forest (FARF). The pipeline of our FARF method, which includes feature extraction, the training stage, and inference stages for SISR, is shown in Fig. 1. In the FARF-based image SR scheme, higher discriminative features are extracted by using the first- and second-order gradients and their magnitudes. Then, the conventional PCA is replaced by the generalized LSH for dimensionality reduction, and the compressed features are used for clustering in the split-nodes on an RF. Finally, the respective regressors at the leaf-nodes are learned by using the original high dimensional features with the GWRR models.

Having introduced the main idea of our paper, the remainder of this paper is organized as follows. In Section 2, we review the related works on SISR, particularly the RF-based approaches and our insights. In Section 3, we introduce the proposed method FARF, including the discriminative feature augmented by the gradient-magnitude filters, the generalized weighted ridge regression (GWRR) model, and the fine-tuned FARF version. In Section 4, we evaluate our FARF scheme on public datasets, and conclusions are given in Section 5.

## 2. IMAGE SUPER-RESOLUTION VIA RANDOM FOREST

**2.1 Image Super-Resolution**

Image SR attempts to achieve an impressive HR quality image from one or a set of LR images via artistic skills, which has been an active research topic for decades in the image restoration area. Generalized SR includes interpolation algorithms, such as the classic bicubic interpolation, and other edge-preserving algorithms [41, 42, 43, 44, 51].

The traditional super-resolution algorithms are based on pixel operations. Intuitively, operating on a "big pixel", i.e. a patch [52], is more effective. Since patch-based algorithms can preserve the local texture structure of an image, various methods based on image patches, such as non-local means [51], self-similarity [31], manifold learning [29], block-matching and 3D filtering (BM3D) [53], sparse representation [28], etc. have been proposed.

The neighbor-embedding (NE) methods [29, 30] are the milestone for patch-based dictionary learning methods. NE learns the mapping between low- and high-resolution patches, with the use of manifold learning. Based on the locally linear embedding (LLE) theory, an LR patch can be represented as a linear combination of its nearest neighbors in a learned dictionary, and its HR counterpart can be approximated as a linear combination of the corresponding HR patches of its LR neighbors, with the same coefficients. Although the NE method is simple and sounds practical, a problem with the method is how to build a feasible patch dictionary. For example, for a patch size of 5×5, with 256 gray levels, it is necessary to have a massive dataset, which has millions of patches, in order to achieve high-quality reconstructed HR patches, if the patches are collected directly from natural scene images. Because of the large dictionary size, it is time consuming to search for a neighbor in such a large dataset.

Other method to reduce the dictionary size is to learn a relatively smaller dictionary with discrete cosine transform (DCT) or wavelet fixed basis, which the adaptiveness is sacrificed. In 2010, Yang *et al*. [28] proposed a sparse prior for dictionary learning. Using sparse coding, image representation can work with a relatively smaller dictionary while keep the adaptiveness by learning the basis from data directly, which opens the era for sparse coding in the image inverse problems.

With the sparse constraint used in the sparse-coding super-resolution (ScSR) framework, an LR patch and its corresponding HR patch can both be reconstructed through two learned coupled dictionaries, with the same coefficients as following:

$$y \approx D_l\alpha, \ x \approx D_h\alpha, \ \alpha \in R^k \ \text{with} \ \|\alpha\|_0 \ll k. \tag{1}$$

where $x$ and $y$ denote an LR patch and its HR counterpart, respectively, and $D_l$ and $D_h$ are the low and high-resolution coupled dictionaries trained jointly from LR and HR patch samples. The value of $\vartheta$ in $\|\alpha\|_\vartheta$ is the sparsity factor of the coefficients $\alpha$. $\|\alpha\|_0$, called the $l^0$-norm, is the non-zero count of the coefficients in $\alpha$. The LR and HR coupled dictionaries are trained jointly with a sparsity constraint, as following:

$$D_h, D_l = \underset{D_h, D_l}{\mathrm{argmin}} \|x - D_h\alpha\|_2^2 + \|y - D_l\alpha\|_2^2 + \lambda\|\alpha\|_0, \tag{2}$$

an LR patch $y$ of an input LR image Y can be formulated in terms of $D_l$ as following:

$$\min\|\alpha\|_0 \ \text{s.t.} \ \|D_l\alpha - y\|_2^2 \leq \varepsilon, \tag{3}$$

or

$$\min\|\alpha\|_0 \ \text{s.t.} \ \|FD_l\alpha - Fy\|_2^2 \leq \varepsilon, \tag{4}$$

where $F$ is a feature-extraction operator on the LR patches, which aims to extract discriminative features from LR patches, rather than using the raw pixel intensity.

Although the $l^0$-norm of α is an ideal regularization term for the sparse constraint, this strong constraint leads to an NP-hard problem in solving the coefficients α. Yang *et al*. [28] relaxed the $l^0$-norm to $l^1$-norm, so as to achieve a feasible solution as following:

$$\min\|\alpha\|_1 \ \text{s.t.} \ \|FD_l\alpha - Fy\|_2^2 \leq \varepsilon, \tag{5}$$

and an equivalent formulation can be achieved by using the Lagrange multiplier,

$$\min_\alpha \|FD_l\alpha - Fy\|_2^2 + \lambda\|\alpha\|_1, \tag{6}$$

where the parameter $\lambda$ balances the sparsity of the solution and the fidelity of the approximation to $y$. As the sparse constraint in [28] is still a bottleneck on training dictionaries considering the computation, an intuitive way to solve it is to relax the constraint again to $l^2$-norm. Meanwhile, the effectiveness of sparsity is challenged [1, 5] by researchers as to whether sparsity or collaborative representation really helps in image classification and restoration. As a natural solution to that, Timofte *et al*. proposed an anchored neighborhood regression (ANR) [2] framework, where there is no sparse constraint in the model. ANR replaces the sparse-decomposition optimization ($l^1$-norm) with a ridge regression (i.e. $l^2$-norm), where the coefficients can be computed offline and each coefficient can be stored as an atom (anchor) in the dictionary. This offline learning can greatly speed-up the prediction stage, and this approach has subsequently led to several variant algorithms.

Timofte *et al*. later extended the ANR approach to the A+ [5]. In A+ [5], the coupled dictionaries are trained from a large pool of training samples (in the order of millions) rather than only from the anchoring atoms, which greatly improves the image quality. After that, more extensions based on ANR and A+ have emerged [1, 33, 34, 35, 36].

However, in the above-mentioned dictionary-learning methods, the complexity of finding those similar patches by comparing an input patch with all the dictionary items has been overlooked. Recently, algorithms using random forest (RF) [2, 5, 7] have achieved state-of-the-art performances, in terms of both accuracy and efficiency for classification and regression tasks. This is mainly due to the use of ensemble learning and sublinear search based on binary trees. Schulter *et al*. [8] adopted random forest and the clustering-regression scheme to learn regressors from the patches in leaf-nodes for SISR. With the same number of regressors, the RF-based algorithm can outperform or achieve comparable performance with A+ and its variants, in terms of accuracy but with less computational complexity.

In recent years, deep learning has achieved promising performances on image super-resolution [37, 38, 39, 40]. In [37, 38], milestone works on image super-resolution based on deep learning were presented, where a convolutional neural network (SRCNN) was proposed to learn an end-to-end mapping between LR and HR images for image super-resolution. Later a scheme with very deep networks for SISR was proposed in [39], where the convergence rate of the deep network is improved by using residual learning and extremely high learning rates. In addition, Ledig *et al.*[40] introduced a generative adversarial network (GAN) based image super-resolution model (SRGAN), where the image perceptual loss function is reformulated as the combination of content loss and adversarial loss. Although deep-learning-based approaches have achieved promising progress on SISR, the heavy computational requirement is still a large burden even though the implementation is accelerated by GPU. This may limit them from those applications without powerful GPU, such as smart mobile terminals.

**2.2 Image Super-Resolution via Random Forest**

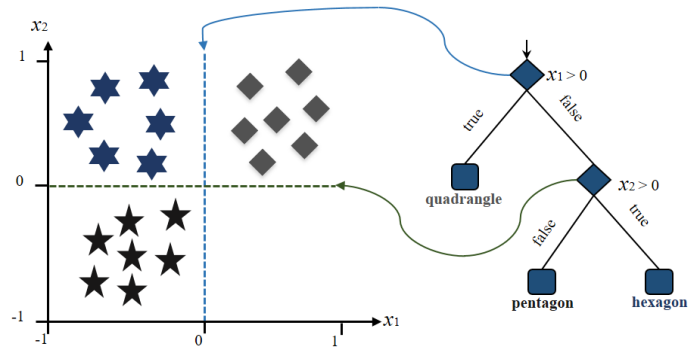

Fig. 2: Random forest for clustering data.

A random forest is an ensemble of binary decision trees $\mathcal{T}^t(x): V \rightarrow R^d$, where $t = 1, \cdots, T$ is the index of the trees, $x$ is a sample from the *m*-d feature space: $V \in R^m$, and $R^d = [0,1]^d$ represents the space of class probability distributions over the label space $Y = \{1, \ldots, d\}$. As shown in Fig. 2, the vertical dotted line forms a hyperplane: $x_1=0$, chosen at the first split node, i.e. the root node, to separate all the training samples, and the horizontal dotted line is the hyperplane: $x_2=0$, for the second split node to cluster all the feature data assigned to this node. This results in separating the three data samples (quadrangle, pentagon and hexagon) into three leaf nodes.

In the inference stage, each decision tree returns a class probability $p_t(y|v)$ for a given test sample $v \in R^m$, and the final class label $y^*$ is then obtained via averaging, as follows:

$$y^* = \arg\max_y \frac{1}{T}\sum_{t=1}^{T} p_t(y|v), \qquad (7)$$

A splitting function $s(v;\Theta)$ is typically parameterized by two values: (i) a feature dimensional index: $\Theta^i \in \{1, \ldots, m\}$, and (ii) a threshold $\Theta^t \in \mathbb{R}$. The splitting function is defined as follows:

$$s(v;\Theta) = \begin{cases} 0, & \text{if } v(\Theta^i) < \Theta^t, \\ 1, & \text{otherwise}, \end{cases} \qquad (8)$$

where the outcome defines to which child node $v$ is routed, and 0 and 1 are the two labels belonging to the left and right child node, respectively. Each node chooses the best splitting function $\Theta^*$ out of a randomly sampled set $\{\Theta^i\}$, and the threshold $\Theta^t$ is determined by optimizing the following function:

$$I = \frac{|L|}{|L|+|R|}H(L) + \frac{|R|}{|L|+|R|}H(R), \tag{9}$$

where $L$ and $R$ are the sets of samples that are routed to the left and right child nodes, respectively, and $|S|$ represents the number of samples in the set $S$. During the training of an RF, the decision trees are provided with a random subset of the training data (i.e. bagging), and are trained independently. Training a single decision tree involves recursively splitting each node, such that the training data in the newly created child node is clustered conforming to class labels. Each tree is grown until a stopping criterion is reached (e.g. the number of samples in a node is less than a threshold or the tree depth reaches a maximum value) and the class probability distributions are estimated in the leaf nodes. After fulfilling one of the stopping criteria, the density model $p(y)$ in each leaf node is estimated by using all the samples falling into the leaf node, which will be used as a prediction of class probabilities in the inference stage. A simple way to estimate the probability distribution function $p(y)$ is by averaging all the samples in the leaf node, and there are many variants, such as fitting a Gaussian distribution, kernel density estimation, etc.

In (9), $H(S)$ is the local score for a set of samples in $S$ ($S$ is either $L$ or $R$), which is usually calculated by entropy, as shown in Eqn. (10), and it can be replaced by variance [8, 18, 21] or by the Gini index [14].

$$H(S) = -\sum_{k=1}^{K}[p(k|S) * \log(p(k|S))], \tag{10}$$

where $K$ is the number of classes, and $p(k|S)$ is the probability for class $k$, which is estimated from the set $S$. For the regression problem, the differential entropy is used, and is defined as,

$$H(q) = \int_y q(y|x) * \log(q(y|x))d_y, \tag{11}$$

where $q(y|x)$ denotes the conditional probability of a target variable given an input sample. Assuming that $q(.,.)$ is of Gaussian distribution, and has only a set of finite samples, the differential entropy can be written as,

$$H_{Gauss}(S) = \frac{K}{2}(1 - \log(2\pi)) + \frac{1}{2}\log(\det(\Sigma_S)), \tag{12}$$

where $\det(\Sigma_S)$ is the determinant of the estimated covariance matrix of the target variables in $S$.

RF-based approaches hold some properties, which make them powerful classifiers as SVM (support vector machine) [10] and AdaBoost (short for "Adaptive Boosting") [13]. Both SVM and AdaBoost work as to approximate the Bayes decision rule – known to be the optimal classifiers – via minimizing a margin-based global loss function.

RF-based image super-resolution (SR), following a recent emerging stream [5, 31] on single-image SR, formulates the SR problem as a clustering-regression problem. These emerging approaches attempt to reconstruct an HR image from patches with the aid of an external database. These methods first decompose an image into patches, then classify the patches into different clusters, and later regressors are trained for all the clusters respectively, which generate mappings from an input LR patch's features to its corresponding HR patch. In the inference stage, an LR image follows the same procedures, such that it is divided into patches and features are extracted from each patch. Then, the patches are classified into different clusters using K-NN [8, 19] or RF [2, 5, 7], and their super-resolved HR patches are computed through regression in the leaf nodes (see Fig. 1). This kind of clustering-regression-based random forest [2, 5, 7] methods has achieved state-of-the-art performance in SISR, both in terms of accuracy and efficiency.

## 3. FEATURE-AUGMENTED RANDOM FOREST

Classification and regression can be regarded as probability problems from the statistical theory. Historical frequentist probability is the probability obtained from the relative frequency in a large number of trials. In contrast, the Bayesian probability is an interpretation of the concept of probability, in which probability is interpreted as an expectation taking the knowledge and personal belief into account. From the Bayesian theory, the posterior probability of a random event is a conditional probability, which can be calculated if the relevant evidence or context is considered. Therefore, the posterior probability is the probability $p(\theta|x)$ of the parameters $\theta$ given the evidence $x$. We denote the probability distribution function of the prior for parameters $\theta$ as $p(\theta)$, and the likelihood as $p(x|\theta)$, which is the probability of $x$ given $\theta$. Then, based on the Bayesian rule, the posterior probability can be defined as follows:

$$p(\theta|x) = \frac{p(x|\theta)p(\theta)}{p(x)}. \tag{13}$$

The posterior probability can be denoted in a memorable form as:

$$Posterior\ probability\ \propto\ Likelihood \times Prior\ probability.$$

Based on the Bayesian framework, the likelihood term and the prior term are both required to be determined in order to solve the inverse problems, and the extracted features are normally worked as prior or likelihood, particularly on some image restoration problems. From this point of view, most research works, from classic feature extractors to deep-learning neural networks, are essentially done under the Bayesian inference framework.

Since SISR is a well-known ill-posed problem, researchers have put their efforts into the priors of the problem with skills from mathematics, computer vision and machine learning. One of the obvious and most studied priors is the edge prior, which can be found in many pioneering works: new edge-directed interpolation (NEDI) [41], soft-decision adaptive interpolation (SAI) [42], directional filtering and data-fusion (DFDF) [43], modified edge-directed interpolation (MEDI) [44], and so on. The edge prior is effective on image processing, and the first and second-order gradients are studied and employed by Yang *et al*. [28] in a pioneering dictionary-learning-based algorithm. However, the effect of edge-based features has not been investigated in depth.

### 3.1 Augmented Features via Gradient Magnitude Filters

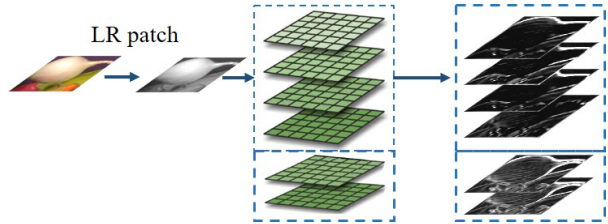

Fig. 3: Features extracted from LR image patches through the first and second-order gradients and gradient magnitude filters, are concatenated to form augmented features with more discriminative

For the clustering and classification problems, feature engineering is a critical research point, and in some cases, the chosen feature may dominate the performance. As shown in Eqn. (6), a feature filter $F$, whose coefficients are computed to fit the most relevant parts in the LR image patches, is employed, and the generated features can achieve more accurate predictions for reconstructing their counterpart HR image patches, as shown in Fig. 3.

Normally it is unstable to directly use pixel intensities as features, which are susceptible to the environmental lighting variations and camera noise. Instead, the differences between the neighboring pixels' intensity values, which are computationally efficient, and are immune to lighting changes and noise, are examined. This type of features can be implemented efficiently through convolutional filters. Typically, the feature filter $F$ can be chosen as a high-pass filter, while in [2, 4, 5, 28], the first and second-order gradient operators are used to generate an up-sampled version from a low-resolution image, then four patches are extracted from the gradient maps at each location, and finally the patches are concatenated to form feature vectors. The four 1-D filters used to extract the derivatives are described in Eqn. (14),

$$\left.\begin{aligned} F_1 &= [-1, 0, 1], \quad F_2 = F_1^T \\ F_3 &= [1, 0, -2, 0, 1], \quad F_4 = F_3^T \end{aligned}\right\}. \tag{14}$$

These features can work well on dictionary-learning-based methods, because when searching a matched patch in a dictionary, the distance is calculated based on the whole feature vectors with the Euclidean distance. However, when training a split node in a decision tree of an RF, only one or a few of the feature dimensions are chosen as candidate features for comparison. Therefore, more discriminative features are required for RF, when compared with dictionary-learning-based methods.

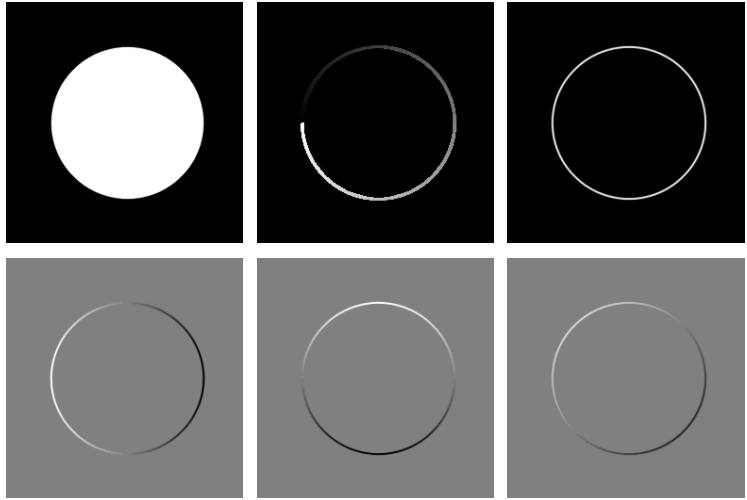

Fig. 4: Visualization of the features from a generated image. Row-1: an original gray image, the *orientation* (*gradient* angle) image, and the *gradient magnitude* image; Row-2: *horizontal gradient* $\partial I/\partial x$, *vertical gradient* $\partial I/\partial y$, and the sum: $(\partial I/\partial x + \partial I/\partial y)$.

The first and second-order *gradients* of an image can provide the directions of edges in a perceptual manner as shown in Fig. 4 and Fig. 5, which can be calculated as Eqn. (15),

$$\nabla I = \left[\frac{\partial I}{\partial x}, \frac{\partial I}{\partial y}\right], \tag{15}$$

where $\partial I/\partial x$ and $\partial I/\partial y$ are the *gradients* in the *x*-axis and *y*-axis directions, respectively, at a given pixel. Meanwhile, the *gradient magnitude* image can provide the edge strength, as described in Eqn. (16). Fig. 4 shows a toy example of a man-made "circle" image, to demonstrate its discriminative property.

$$\|\nabla I\| = \sqrt{(\frac{\partial I}{\partial x})^2 + (\frac{\partial I}{\partial y})^2}. \tag{16}$$

With a natural image shown in Fig. 5, it can be observed that the *gradient magnitude* image has more detailed textures than the *gradient* images ($\partial I/\partial x$ and $\partial I/\partial y$), as well as the sum of the horizontal *gradient* and vertical *gradient* image, i.e. $\partial I/\partial x + \partial I/\partial y$, perceptually. An explanation for this phenomenon is that non-linear features are usually more discriminative. Thus, in our work, all the first and second-order *gradients,* and *gradient magnitude* are employed, and are concatenated to form more discriminative, augmented features.

On the other hand, the image *orientation* (gradient angle) is defined by the following formulation,

$$\angle \nabla I = \arctan(\partial y / \partial x), \qquad (17)$$

where atan( ) is the gradient orientation, with a value between -90° and 90°. As shown in Eqn. (17), when the value of $\partial x$ is equal to 0 or close to 0, the value of $\angle \nabla$ becomes infinitely large and unstable, i.e., different $\partial y$ will result in approximately the same $\angle \nabla$ value.

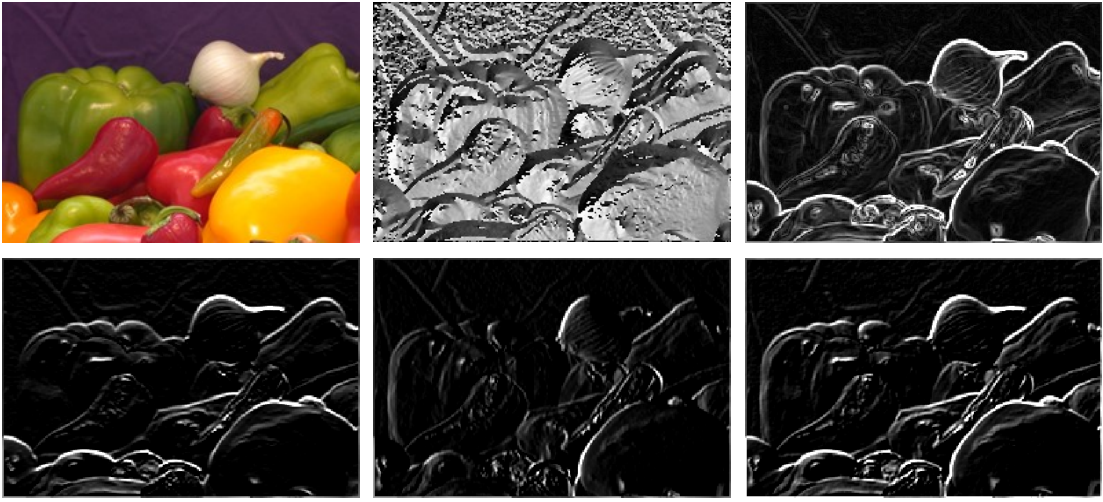

Fig. 5: Visualization of the features from a natural image. Row-1: original color image, image *gradient orientation* and image *gradient magnitude*; Row-2: *horizontal gradient* $\partial I/\partial x$, v*ertical gradient* $\partial I/\partial y$ and the sum: $(\partial I/\partial x + \partial I/\partial y)$.

Based on this analysis, we only use the two *gradient magnitude* filters derived from the four gradient filters [28] to generate the augmented features. Experiments validate that the use of the augmented features can improve the conventional RF algorithm [8] to achieve a performance gain of more than 0.1dB, which is a remarkable improvement, with the same setting and parameters.

### 3.2 Fine-grained Features for Regression

The inference stage of the RF-based image super-resolution process is similar to the content-based image retrieval (CBIR) framework, as shown in Fig. 1. The general approximated nearest neighbor (ANN) search framework [46, 47] is an efficient strategy for large-scale image retrieval, which mainly consists of 4 parts: (1) extracting compact features (e.g., locality-sensitive Hashing (LSH) [48] feature) for a query image; (2) coarse-level search using Hamming distance to measure the similarity between binary compact Hash features, then narrow the search scope into a smaller candidate group; (3) fine-level search by using Euclidean distance to measure the similarity between their corresponding feature vectors; and (4) finding the object in the smaller candidate group that is the nearest one to the query image.

In the inference stage of conventional RF-based SISR, PCA projection is worked as a Hash-like function to compress the feature dimension for decreasing the search range, which can speed up the searching as the coarse-level search in a CBIR framework, but the impact of using PCA on feature dimensionality reduction has been overlooked in previous works [1, 2, 4, 5, 8, 28]. Inspired by the fine-level search using augmented features in CBIR frameworks, the high dimensional features in the leaf nodes in an RF can further improve the prediction accuracy in the regression step, which has not been studied previously. Consequently, we use the original features, rather than PCA or the LSH compressed features, to perform ridge regression in the leaf nodes. Experimental results show that the new RF scheme can greatly improve the quality of super-resolved images, by using this augmented feature. Another explanation for this is that the regression problems can benefit more from higher dimensional features than classification problems.

Based on the observation that the original edge-like features are used for the final regressors in the leaf nodes and the compressed features (either produced by PCA or LSH) are used for clustering in the split nodes, a new clustering-regression-based SISR approach can be designed as shown in Fig. 6. In this new scheme, the original-compressed coupled feature sets are worked for different purposes at different stages, i.e., the original edge features are used for regression in the leaf nodes, and the compressed features derived from the LSH-like functions are employed for node splitting (clustering) in the training stage, and node searching in the inference stage in the split nodes.

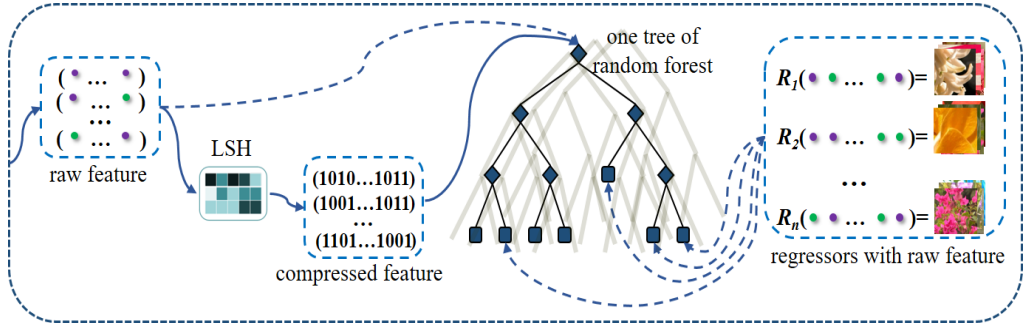

Fig. 6: Augmented features for regressors and the LSH compressed features for searching in a random forest

In the new scheme, we unify the research of LSH-based SISR and image retrieval (CBIR) [46, 47]. In brief, the new achievement on unsupervised LSH can be evaluated not only in CBIR systems, but also in the clustering-regression RF-based SISR methods. Moreover, as evidence from [56], proper unsupervised LSH models, e.g., iterative quantization (ITQ) [57] used for feature dimension reduction instead of PCA, can reduce the damage on the image structure. This can further improve the super-resolved image quality. Different from [56] using an ITQ-like algorithm to rotate the original features into a new feature space, with the use of the proposed original-compressed coupled feature sets, any unsupervised LSH generated features can directly be employed.

### 3.3 Generalized Weighted Ridge Regression Model

In this sub-section, we further analyze the ridge regression employed in the RF leaf nodes. The anchored neighborhood regression (ANR) [2] model relaxes the $l^1$-norm in Eqn. (6) to the $l^2$-norm constraint, with least-squares minimization as the following equation,

$$\min_{\alpha}\|FD_l\alpha - Fy\|_2^2 + \lambda\|\alpha\|_2, \tag{18}$$

Based on the ridge regression [16] theory, this $l^2$-norm constrained least square regression regularized problem has a closed-form solution, according to the Tikhonov regularization theory, as follows:

$$\alpha = (D_l^T D_l + \lambda I)^{-1} D_l^T F y. \tag{19}$$

With the assumption in [28], where HR patches and their counterpart LR patches share the same reconstructed coefficient $\alpha$, i.e. $x = D_h \alpha$, from Eqn. (19) we have

$$x = D_h (D_l^T D_l + \lambda I)^{-1} D_l^T F y. \tag{20}$$

If we define $P_G$ as a pre-calculated projection matrix, as follows,

$$P_G = D_h (D_l^T D_l + \lambda I)^{-1} D_l^T, \tag{21}$$

then the HR patches can be reconstructed with $x = P_G F y$.

Having studied the model in Eqn. (18), the authors in [1] argued that different weights should be given to different atoms when reconstructing an HR patch so as to emphasize the similarity to the anchor atom. Based on this idea, [1] proposed a weighted collaborative representation (WCR) model by generalizing the normal collaborative representation (CR) model in the ANR,

$$\min_\alpha \|F D_l \alpha - F y\|_2^2 + \|\lambda_{WCR} \alpha\|_2, \tag{22}$$

where $\lambda_{WCR}$ is a diagonal weight matrix, in which the non-zero entries are proportional to the similarities between the atoms and the anchor atom.

Same as the ANR model, a new closed-form solution can be computed offline through the following equation,

$$\alpha^* = (D_l^T D_l + \lambda_{WCR})^{-1} D_l^T F y, \tag{23}$$

and the new projection matrix can be derived as

$$P_G^* = D_h (D_l^T D_l + \lambda_{WCR})^{-1} D_l^T. \tag{24}$$

The WCR model further improves the ANR/A+ model in terms of image quality, while keeping the same level of computation. In [9], the local geometry prior of the data sub-space is used. However, all the weighted ridge regression models [1, 9] are constructed based on an existing dictionary, e.g., Zeyde *et al.* [4] used K-SVD to train a sparse-coding-based dictionary with 1024 items. This limits the models to collect samples in a smaller sub-space when constructing linear regressors based on existing anchor points.

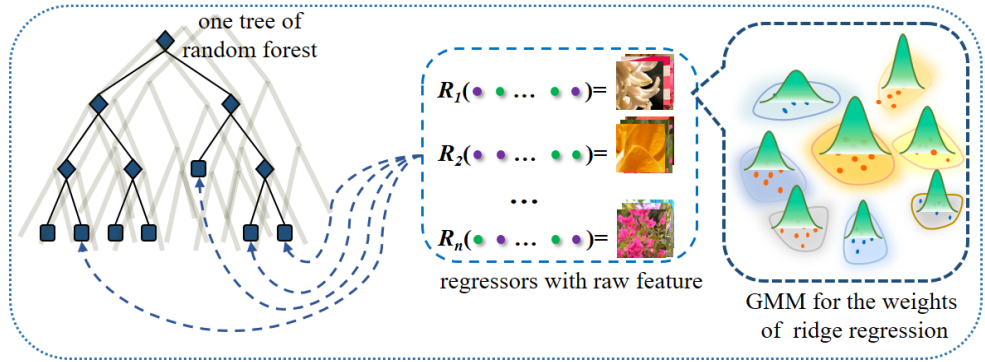

Fig. 7: Gaussian mixture model (GMM) is used to generate the weights for weighted ridge regression, and the weight of each entry lies on its belonging cluster's weight and its weight in the belonging cluster.

When training the regressors in an RF, there is no existing anchor point in the clustered groups of the leaf nodes, similar to the previous models [1, 9]. A solution to mentioned problem is inspired from the work on image classification using locality-constrained linear coding (LLC) [49], where Gaussian mixture model (GMM) is used to describe the locality-constrained affine subspace coding (LASC) [50]. We employ GMM to construct the data distribution in the sub-space for each leaf node, which derives the weights of all the entries in the ridge regression models. Through the derived weights, we can obtain a generalized weighted ridge regression (GWRR) model for ridge regression. The new projection matrix is given as follows:

$$P_G^* = D_h(D_l^T D_l + \lambda_{GWRR})^{-1} D_l^T, \tag{25}$$

where $\lambda_{GWRR}$ is a diagonal weight matrix, and the weight of each diagonal entry is related to its belonging cluster's weight and its local weight in its belongingwhi cluster, as illustrated in the right part of Fig. 7. Obviously, a query entry falling into a bigger cluster and closer to the center of the belonging cluster achieves a larger weight. In a rough form, the diagonal weight matrix $\lambda_{GWRR}$ is given as follows:

$$\lambda_{GWRR} = \text{diag}\{[w_1; w_2; \dots w_i; \dots; w_m]\}, \ w_i \propto C_i^k \times d_i^k, \ k = (1, \dots, K), \tag{26}$$

where $w_i$ is the weight of the $i^{th}$ entry, $m$ is number of samples in the leaf nodes, $C_i^k$ is the $k^{th}$ cluster's weight for the $i^{th}$ entry, $d_i^k$ is the $i^{th}$ entry's local weight in the $k^{th}$ cluster, which is approximated with the inverse value of the distance to the center of the belonging cluster, and $K$ is the number of clusters generated by the GMM model for a leaf node.

Experimental results in Table-1 show that the proposed GWRR model can achieve the same level of performance as WCR [1], and obtain 0.2dB gain more than the ANR [1] model.

| *Images* | baboon | baby | bird | butterfly | foreman | head | lenna | woman | *Average* |
|---|---|---|---|---|---|---|---|---|---|
| ANR | 23.52 | 35.06 | 34.44 | 25.74 | 32.92 | 33.54 | 32.92 | 30.17 | 31.04 |
| WCR | 23.55 | 35.09 | 34.75 | 26.18 | 33.51 | 33.61 | 33.16 | 30.42 | **31.28** |
| GWRR | 23.54 | 35.09 | 34.74 | 26.13 | 33.46 | 33.58 | 33.12 | 30.38 | 31.25 |

Table-1: Performances of ANR [1], WCR [1], and the proposed GWRR, in terms of PSNR (dB) with an upscale factor (×3) on some public standard test images

Note that when the number of samples in a leaf node becomes bigger, the performance of the GWRR model will achieve less advantage than the normal regression model, because the higher weights will be averaged by a large number of other samples. Theoretically, the regression of a leaf node can benefit from the GWRR model, particularly when there are a few samples falling into the leaf node.

### 3.4 Initial Estimation with Iterative Back Projection

Generally speaking, SISR is a low-level computer vision task, which attempts to restore an HR image $\mathcal{X}$ from a single input LR image $\mathcal{Y}$. A mathematical model for image degradation can be formulated as follows:

$$\mathcal{Y} = (\mathcal{X} * \mathcal{B}) \downarrow s, \tag{27}$$

where $\mathcal{B}$ is a low-pass (blur) filter and $\downarrow s$ denotes the down-sampling operator with $s$ factor. Based on a given LR image $\mathcal{Y}$, how to achieve an approximated HR image $\hat{\mathcal{X}}$ is a classic inverse problem, which requires priors based on the Bayesian theory.

Irani and Peleg [54] firstly proposed an iterative back projection (IBP) method for SR reconstruction, and IBP is the most effective way to obtain an HR image when comparing it with other SR methods. In the IBP method, the reconstruction error of an estimated LR image $\hat{\mathcal{Y}}$ is the difference between the input LR $\mathcal{Y}$ and the synthesized image $\hat{\mathcal{Y}}$ generated from the estimated HR image $\hat{\mathcal{X}}$ as follows:

$$e(\hat{\mathcal{Y}}) = \mathcal{Y} - \hat{\mathcal{Y}} = \mathcal{Y} - (\hat{\mathcal{X}} * \mathcal{B}) \downarrow s. \tag{28}$$

IBP is an efficient approach to obtain the HR image by minimizing the reconstruction error defined by Eqn. (28). For the IBP approach on SISR, the updating procedure can be summarized as the following two steps, performed iteratively:

- Compute the reconstruction error $e(\hat{\mathcal{X}})$ with the following equation:

$$e(\hat{\mathcal{X}}) = e(\hat{\mathcal{Y}}) \uparrow s * p, \tag{29}$$

where $\uparrow$ is the up-sampling operator and $p$ is a constant back-projection kernel to approximate the inverse operation of the low-pass filter $\mathcal{B}$.

- Update the estimating HR image $\hat{\mathcal{X}}$ by back-projecting errors as follows:

$$\hat{\mathcal{X}}^{t+1} = \hat{\mathcal{X}}^t + e(\hat{\mathcal{X}}^t), \tag{30}$$

where $\hat{\mathcal{X}}^t$ is the estimated HR image at the $t$-th iteration.

Most learning-based algorithms [1, 2, 4, 5] follow the milestone work in [28], which uses the coarse estimation firstly obtained via bicubic interpolation. As we know, the classic IBP algorithm is an efficient way to obtain high-quality up-scaled images, but it will inevitably produce artifacts (such as ringing, jaggy effects, and noise) at the output, because the kernel operator $p$ in Eqn. (29) is hard to estimate accurately. That is the reason why algorithms with IBP need an additional denoising process [51, 54, 58]. However, the sparse-constraint-based approach [28] does not have this denoising capability.

As the $l^2$-norm constraint-based ridge regression has the denoising effect, due to its averaging-like process, this means that the ridge regression-based RF scheme has the denoise capability intrinsically. Based on this observation, we obtain the coarse estimation of an HR image $\hat{\mathcal{X}}$ by applying IBP to the corresponding input LR image $\mathcal{Y}$. Experimental results in Table-2 and Table-3 validate that using IBP, instead of bicubic, to obtain the initial coarse estimation can help the RF-based SR method obtain a remarkable improvement.

### 3.5 Fine-Tuning with Proper Trees in Random Forest

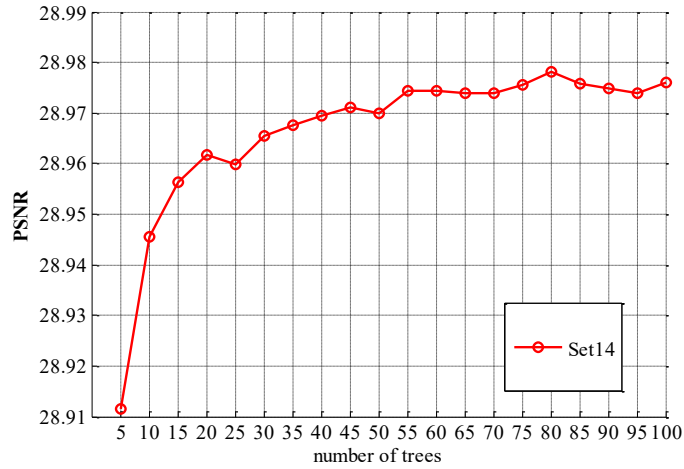

Fig. 8: The image super-resolution quality (i.e., measured by PSNR) with different numbers of trees in a random forest for super-resolution (3x) experiments on Set14. The number of trees = 45 gives a better trade-off between efficiency and complexity.

As the number of trees is an important parameter in RF-based approaches, we plot the performance with respect to the number of trees. As shown in Fig. 8, the performance of the RF-based image super-resolution method increases as expected, but the increment becomes relatively smaller after a certain number of trees are used. The experimental results in Fig. 8 are obtained on the Set14 dataset, and 2 million samples from the dataset are used for all training stages. It shows that using 45 trees is an optimal number, as a trade-off between performance and computational cost. Therefore, we set the number of trees for the proposed FARF method at 45, and our method with this number is denoted as FARF*. The performances of our methods, and other methods, are tabulated in Table-2 and Table-3. We also compare our methods with a recently proposed deep-learning-based algorithm, SRCNN algorithm [37, 38], and our methods outperform it in some cases.

### 3.6 Algorithm Workflow

The training and inference stages of the proposed FARF algorithm are described in Algorithm 1 and Algorithm 2, respectively. To help the readers understand our paper, the source code of our algorithm will be available at: https://github.com/HarleyHK/FARF, for reference.

---

**Algorithm 1: Training Stage of FARF based Image Super-Resolution:**

**Input:** $\{y^i, x^i\}_{i=1}^{N}$: training LR-HR patch pairs;

**Output:** the trained random forest $\mathcal{T}$ with regressors $\mathcal{R} = (\mathcal{R}_1, \dots)$, the LSH model: $\mathcal{M}_{LSH}$;

**1:** Upscale the input LR patch images as initial coarse estimations using IBP;  $\Rightarrow$ {Eqn. (29, 30)}

**2:** Obtain discriminative features calculated from patches by the first-order, second-order (horizontal and vertical) gradients, and gradient magnitudes for up-scaled coarse versions;  $\Rightarrow$ {Eqn. (15, 16)}

**3:** Conduct LSH on raw features to obtain compressed features, at the same time obtain the trained LSH projection model $\mathcal{M}_{LSH}$;

**4:** Train a random forest with the compressed features via the LSH model $\mathcal{M}_{LSH}$;

**5:** Train the weighted ridge regressors $\mathcal{R}$ by the GWRR models in leaf nodes;  $\Rightarrow$ {Eqn. (25)}

**6:** Save the random forest $\mathcal{T}$ with ridge regressors $\mathcal{R}$, and the trained LSH model: $\mathcal{M}_{LSH}$.

---

**Algorithm 2: Inference Stage of FARF based Image Super-Resolution:**

**Input:** testing LR image $\mathcal{Y}$, the trained random forest $\mathcal{T}$ with ridge regressors $\mathcal{R} = (\mathcal{R}_1, \dots)$, the trained LSH model: $\mathcal{M}_{LSH}$;

**Output:** super-solved image $\hat{\mathcal{X}}$;

**1:** Upscale the patches from LR $\mathcal{Y}$ to form an initial coarse estimation by IBP;  $\Rightarrow$ {Eqn. (29, 30)}

**2:** Compute the discriminative features for all the patches;  $\Rightarrow$ {Eqn. (15, 16)}

**3:** Compute the compressed feature via the LSH model $\mathcal{M}_{LSH}$;

**4:** For each patch, using the compressed feature to search the leaf nodes to obtain its corresponding regressor from the trained random forest $\mathcal{T}$;

**5:** Get the super-resolved image $\hat{\mathcal{X}}$ through all the super-solved patches by weighted ridge regressors $\mathcal{R}$ in leaf nodes.  $\Rightarrow$ {Eqn. (22)}

---

### 4. EXPERIMENTS

In this section, we evaluate our algorithm on standard super-resolution benchmarks Set 5, Set14 and B100 [20], and compare it with some state-of-the-art methods. They are *bicubic* interpolation, adjusted anchored neighborhood regression (A+) [5], standard RF [8], alternating regression forests (ARF) [8], and the convolutional neural-network-based image super-resolution (SRCNN) [37, 38], as listed in Table-2 and Table-3. We set the same parameters for all the RF-based algorithms, i.e., the number of trees in an RF is 10, and the maximum depth of each tree is 15. We use the same set of training images (91 images) for all the algorithms, as previous works [2, 4, 5, 8] do. RF+ means a normal RF-based algorithm added with the two *gradient magnitudes* augmented features, and RF# is the normal RF-based algorithm, where the original raw features, instead of using the PCA compressed features, are used to learn the regressors in leaf nodes. FARF denotes our proposed feature-augmented RF scheme, which combines RF+ and RF# by adding the gradient magnitude features and using the original raw features for regression. FARF* is a further refined version of FARF, by performing further fine-tuning: (1) using the superior, unsupervised LSH projection, instead of PCA for dimensionality reduction, (2) employing IBP, instead of the traditional *bicubic* interpolation algorithm, to obtain the initial coarse estimation in the inference stage, (3) setting the proper number of trees (e.g., 45) for training an RF.

| dataset | # | bicubic | A+ | RF | ARF | RF+ | RF# | FARF- | FARF | FARF* | SRCNN |
|---|---|---|---|---|---|---|---|---|---|---|---|
| **Set5** | ×2 | 33.66 | 36.55 | 36.52 | 36.65 | 36.67 | 36.63 | 36.68 | 36.78 | ***36.81*** | 36.66 |
| | ×3 | 30.39 | 32.59 | 32.44 | 32.53 | 32.56 | 32.53 | 32.62 | 32.73 | ***32.78*** | 32.75 |
| | ×4 | 28.42 | 30.29 | 30.10 | 30.17 | 30.18 | 30.22 | 30.27 | 30.39 | *30.45* | ***30.48*** |
| **Set14** | ×2 | 30.23 | 32.28 | 32.26 | 32.33 | 32.37 | 32.32 | 32.37 | 32.41 | ***32.45*** | 32.42 |
| | ×3 | 27.54 | 29.13 | 29.04 | 29.10 | 29.17 | 29.11 | 29.17 | 29.23 | ***29.29*** | 29.28 |
| | ×4 | 26.00 | 27.33 | 27.22 | 27.28 | 27.31 | 27.29 | 27.36 | 27.45 | *27.48* | ***27.49*** |
| **B100** | ×2 | 29.32 | 30.78 | 31.13 | 31.21 | 31.22 | 31.23 | 31.34 | 31.35 | ***31.38*** | 31.36 |
| | ×3 | 27.15 | 28.18 | 28.21 | 28.26 | 28.27 | 28.26 | 28.30 | 28.35 | 28.38 | ***28.41*** |
| | ×4 | 25.92 | 26.77 | 26.74 | 26.77 | 26.78 | 26.79 | 26.83 | 26.88 | ***26.91*** | 26.90 |

Table-2: Results of the proposed method compared with state-of-the-art works on 3 datasets
in terms of PSNR (dB) using three different magnification factors (#) (×2, ×3, ×4).

Table-2 summarizes the performances of our proposed algorithm on the 3 datasets, in terms of the average peak signal to noise ratio (PSNR) scores, with different magnification factors (×2, ×3, ×4). Table-3 gives more details of the results on some images from the Set5 dataset, with magnification factor ×3. As the results have shown based on the 3 datasets, our proposed algorithm FARF outperforms A+ and ARF for all the magnification factors.

| **Set5**(×3) | bicubic | Zeyde | A+ | RF | ARF | FARF- | FARF | FARF* | SRCNN |
|---|---|---|---|---|---|---|---|---|---|
| baby | 33.91 | 35.13 | 35.23 | 35.25 | 35.15 | 35.20 | 35.34 | ***35.37*** | 35.25 |
| bird | 32.58 | 34.62 | 35.53 | 35.23 | 35.31 | 35.39 | 35.53 | ***35.54*** | 35.47 |
| butterfly | 24.04 | 25.93 | 27.13 | 27.00 | 27.39 | 27.65 | 27.68 | 27.82 | ***27.95*** |
| head | 32.88 | 33.61 | 33.82 | 33.73 | 33.73 | 33.75 | 33.84 | ***33.85*** | 33.71 |
| woman | 28.56 | 30.32 | 31.24 | 30.98 | 31.08 | 31.11 | 31.27 | *31.34* | ***31.37*** |
| ***average*** | 30.39 | 31.92 | 32.59 | 32.44 | 32.53 | 32.62 | 32.73 | ***32.78*** | 32.75 |

Table-3: Results of the proposed method compared with state-of-the-arts methods on 3 datasets in terms of PSNR (dB) with
magnification factors (×3) on dataset Set5.

The objective quality metric, PSNR, in Table-2 also shows that the fine-tuned FARF, i.e. FARF*, can further improve the image quality, which is comparable to recently proposed state-of-the-art deep-learning-based algorithms, such as SRCNN [37, 38].

Comparing our proposed FARF algorithm to other methods, the improved visual quality of our results is obvious, as shown in Fig. 9. This shows that our method can produce more details, particularly on some texture-rich regions.

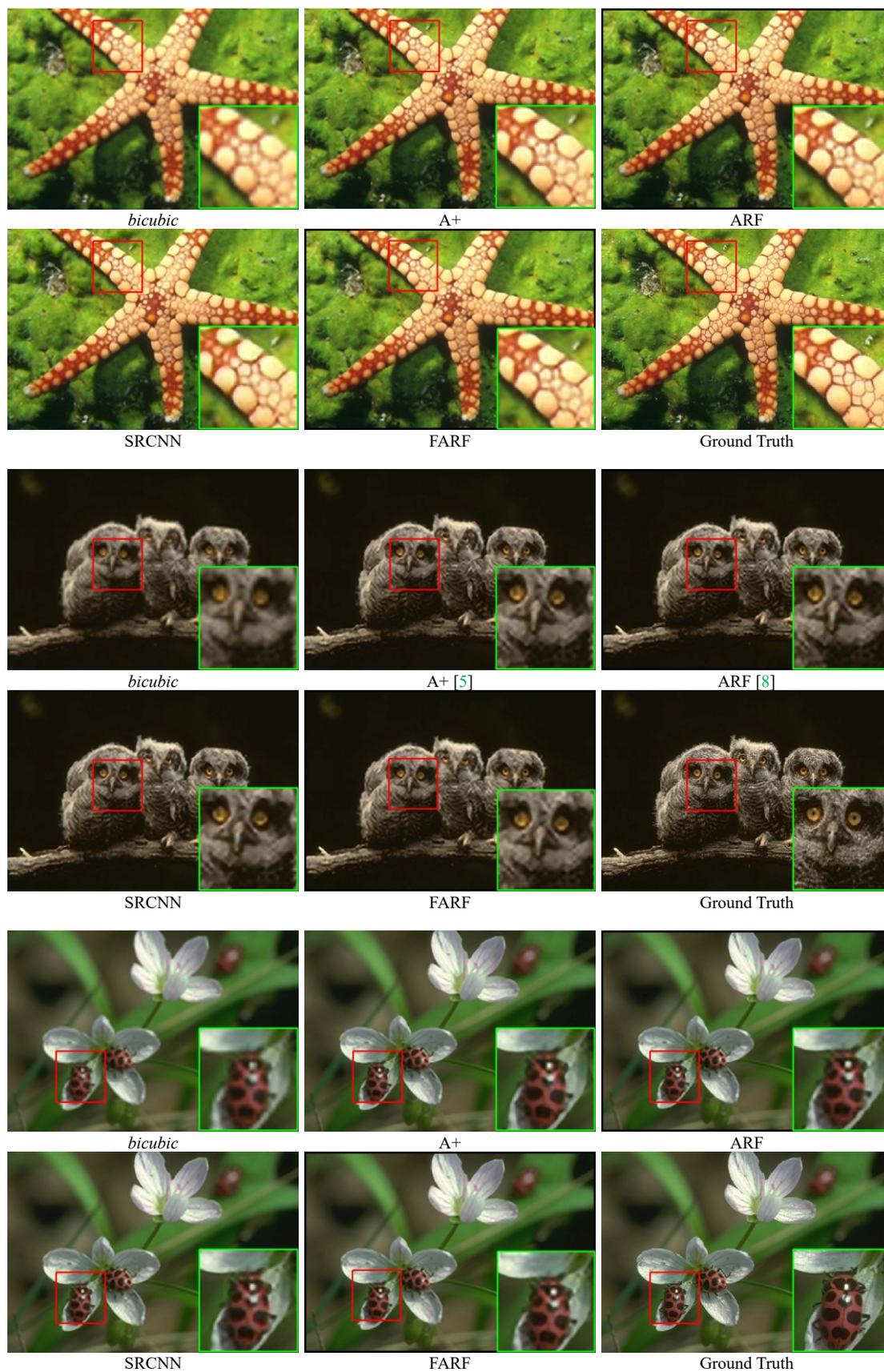

Fig. 9: Super-resolution (×3) images from B100, *bicubic*, A+ (ACCV-2014) [5], ARF (CVPR-2015) [8], SRCNN (PAMI-2016) [38], our proposed algorithm FARF, and ground truth. The results show that our FARF algorithm can produce more details and its performance is comparable to a recent state-of-the-art deep-learning method [38].
16

5. CONCLUSIONS

This paper presents a feature-augmented random forest (FARF) scheme for the single image super-resolution (SISR) task by augmenting features and redesigning the inner structure of a random forest (RF), with different feature recipes at different stages, where the compressed features are used for clustering in the split nodes and the original features are used for regression in the leaf nodes. The contributions of this paper are threefold: (1) the more discriminative gradient magnitude-based augmented features are proposed for clustering on split nodes and regression on leaf nodes; (2) By extending principal component analysis (PCA) to a generalized unsupervised locality-sensitive hashing (LSH) model for dimensionality reduction, we lay out an original compressed coupled feature set for tackling the clustering-regression tasks, which unify SISR and content-based image retrieval (CBIR) for LSH evaluation; and (3) we have extended WCR model to a generalized GWRR model for ridge regression. The proposed FAFR scheme can achieve highly competitive quality results, e.g., obtaining about a 0.3dB gain in PSNR, on average, when compared to conventional RF-based super-resolution approaches. Furthermore, a fine-tuned version of our proposed FARF approach is provided, whose performance is comparable to recent state-of-the-art deep-learning-based algorithms.